\documentclass{article}
\usepackage{spconf,amsmath,graphicx}
\usepackage{acronym}
\usepackage{amssymb}
\usepackage{booktabs,multirow}
\usepackage[normalem]{ulem}
\usepackage{xcolor}

\newcommand{\todo}[1]{}
\newcommand{\editpostreview}[2]{#2}
\newcommand{\editpostsubmission}[2]{#2}

\newcommand{\R}{\mathbb{R}}

\newcommand{\AnyModel}{\ensuremath{\mathbf{\Theta}}}
\newcommand{\AnyParameter}[1]{\ensuremath{\Theta_{#1}}}
\newcommand{\Importance}[1]{\ensuremath{I_{#1}}}
\newcommand{\Likelihood}{\ensuremath{\mathcal{L}}}
\newcommand{\Data}{\ensuremath{\mathbf{X}}}
\newcommand{\Gradient}[1]{\ensuremath{\nabla_{#1}}}
\newcommand{\reldiff}[1]{\emph{#1}}

\newacro{ASR}{automatic speech recognition}
\newacro{BPE}{byte-pair encoding}
\newacro{CPU}{central processing unit}
\newacro{FF}{feed-forward}
\newacro{FLOP}{floating point operation}
\newacro{GPU}{graphic processing unit}
\newacro{HMM}{hidden Markov model}
\newacro{DD}{data-driven}
\newacro{LM}{language model}
\newacro{MD}{magnitude-driven}
\newacro{NE}{named entity}
\newacro{NER}{named-entity recognition}
\newacro{NNLM}{neural network language model}
\newacro{NLP}{natural language processing}
\newacro{PPL}{perplexity}
\newacro{RELU}{rectified linear units}
\newacro{SNEER}{slot named-entity error rate}
\newacro{SVD}{singular value decomposition}
\newacro{TT}{Transformer Transducer}
\newacro{TTS}{text-to-speech}
\newacro{VA}{voice assistant}
\newacro{WER}{word error rate}
\newacro{MAPSSWE}{matched pairs sentence-segment word error}

\title{Neural Language Model Pruning for Automatic Speech Recognition}

\name{Leonardo Emili$^{1,}$\sthanks{Work done during an internship at Apple.}\sthanks{Equal contribution.}, Thiago Fraga-Silva$^{2,}\footnotemark[2]$, Ernest Pusateri$^{2}$, Markus Nußbaum-Thom$^{2}$, Youssef Oualil$^{2}$}
\address{$^{1}$Sapienza University of Rome, $^{2}$Apple
\\[1ex]
{\normalsize \texttt{leonardo.emili@uniroma1.it}}
\\
{\normalsize \texttt{\{tfragadasilva,epusateri,mnussbaumthom,youalil\}@apple.com}}}

\begin{document}
\maketitle

\begin{abstract}
We study model pruning methods applied to Transformer-based \aclp{NNLM} for \acl{ASR}.
We explore three aspects of the pruning framework, namely \emph{criterion}, \emph{method} and \emph{scheduler},
analyzing their contribution in terms of accuracy and inference speed.
To the best of our knowledge, such in-depth analyses on large-scale recognition systems
has not been reported in the literature.
In addition, we propose a variant of low-rank approximation suitable for incrementally
compressing models, and delivering multiple models with varied target sizes.
Among other results, we show that a) data-driven pruning outperforms magnitude-driven in several scenarios;
b) incremental pruning achieves higher accuracy compared to one-shot pruning, especially when targeting smaller sizes;
and c) low-rank approximation presents the best trade-off between size reduction and inference speed-up for moderate compression.
\end{abstract}
\section{Introduction}

Recent literature has shown that increasing the size of \acp{NNLM} can improve accuracy for several \acl{NLP} tasks \cite{devlin-etal-2019-bert-short, Brown2020, raffel2020t5-short}.
\editpostreview{On the other hand, employing}{Employing} large models on applications with memory and complexity constraints can be challenging. For such cases, the default strategy is to estimate small footprint models that meet the given constraints. However, better accuracy levels can be obtained by starting from a large model and shrinking it to the target size \cite{murray2015auto, suau2020filter, zhen2021sparsification}.
Popular model size reduction strategies include, among others, knowledge distillation \cite{Gou_2021}, weight quantization \cite{gholamisurvey2022}, low rank layer factorization \cite{kumar2016survey} or model pruning \cite{zhu2018toprune, liang2021pruning}.

Model pruning has long been investigated as a way to effectively compress large models
\cite{lecun1986optimal}.
Defining which parameters to remove is one \editpostreview{}{of the} aspects to consider.
One of the first criteria proposed is to remove low-\textit{magnitude} parameters, following the intuition that parameter relevance correlates well with their magnitude \cite{mao-etal-2020-ladabert-short,gordon2020}. Other works take the \textit{data} distribution into account to define the parameter importance, making such methods suitable for joint pruning and task-adaptation \cite{lecun1986optimal, mozer1988skeletonization}.

Another aspect to consider is the pruning method. \textit{Unstructured pruning}, or sparsification, works by removing a certain number of neuron connections \editpostreview{}{presumed to be redundant}.
\editpostreview{Removed connections are presumed to be redundant.}{}
Such process yields regularization effects similar to dropout \cite{srivastava2014dropout-short}.
Instead of removing individual connections, \textit{structured pruning} enforces the presence of block sparsity patterns, \editpostreview{. Such methods provide}{which can improve} memory consumption and latency \cite{liu-etal-2021-ebert-short}.
\textit{Low-rank approximation} methods \cite{eckart1936approximation},
on the other hand, rely on factorizing the layer projections into the multiplication of two or more smaller matrices. Then, the inner channels that carry less information to reconstruct the original matrices are removed.

The pruning scheduling is another aspect to take into account.
Usually, the straightforward choice is to remove parameters in a single step (hereafter \textit{one-shot}). More recently, \cite{zhu2018toprune} showed that inducing sparsity \textit{incrementally} while training can let the model recover from pruning losses, yielding better classification rates. 

In this paper, we investigate different choices involved in model pruning, and to what extent generic pruning assumptions are true in the context of language modeling for \ac{ASR}.
Our main contributions are:

\begin{itemize}
  \item Perform an in-depth evaluation of the aforementioned aspects of pruning, namely:
    \textit{pruning criteria} (magnitude and data driven);
    \textit{pruning methods} (unstructured, structured and pruning factorized layers);
    and \textit{pruning scheduling} (one-shot and incremental).
  \item Propose a variant of low-ranking approximation suitable for training pipelines delivering multiple models with varied sizes.
  \item Benchmark pruning approaches applied to \editpostsubmission{large-scale} language models used in state-of-the-art \ac{ASR} systems
  in terms of accuracy, model size and inference \editpostsubmission{speed}{complexity}.
\end{itemize}

\section{Related work}

Previous work analyzed ways to reduce the footprint of a pre-trained model, such as pruning the redundant connections~\cite{zhen2021sparsification, zhu2018toprune, mozer1988skeletonization, qiu2020blockwise-short, molchanov2019importance} or the $n$-grams with the smallest impact on training set perplexity~\cite{DBLP:conf/icassp/GondalaVPTG21}, transferring the acquired knowledge to a smaller student model~\cite{https://doi.org/10.48550/arxiv.2101.04731-short,
https://doi.org/10.48550/arxiv.1603.05691,DBLP:journals/corr/HintonVD15}, optimizing the numerical representation range via quantization~\cite{9251854-short, https://doi.org/10.48550/arxiv.1612.01064},
and applying low-rank factorization \editpostreview{methods}{techniques} to reduce the computational overhead of streamable \ac{ASR} systems~\cite{6638949-short}.

A more recent trend uses the correlation with the training data to compress the model while adapting on specific domains~\cite{suau2020filter, hsu2022language}. 
Other \editpostreview{methods}{approaches} employ similar ideas to create more robust networks by favouring an equal contribution of all the existing parameters instead of pruning them~\cite{xu2022importance, liang2022noparameters}.

Although often applied individually, model size reduction techniques can be combined to yield better results~\cite{ganesh-etal-2021-compressing-short, https://doi.org/10.48550/arxiv.2011.09058, wang2019structured}.
In Section~\ref{par:factorized_layers}, we propose a technique to efficiently prune a factorized architecture, similar to~\cite{DBLP:conf/emnlp/WangWL20}, that can be iterated to \emph{incrementally} generate smaller dense models that meet multiple resource constraints.

\section{The pruning framework} \label{sec:research_questions}
This section describes key challenges involved with model pruning that are investigated in this paper.
Hereby, \textit{sparsity} is considered the ratio between the number of removed connections compared to the initial fully-connected network. Generally, the more sparse, the smaller the model.

\subsection{Pruning criteria}
\label{sec:pruning_criteria}

The \textit{pruning criterion} defines which parameters should be removed at each pruning step.
Given a model with $N$ parameters $\AnyModel = \{\AnyParameter{i}\}$, $i \in [1 .. N]$, we define the 
importance score \Importance{}(\AnyParameter{i}) of the weight \AnyParameter{i} as the driving criterion for pruning.
Pruning frameworks seek to remove parameters with the lowest importance scores, which presumably contribute less to model predictions. 

\Ac{MD} methods use the magnitude of weights as importance scores: 

\begin{equation}
    \label{eqn:mb_importance}
    \Importance{MD}(\AnyParameter{i}) = |\AnyParameter{i}| .
\end{equation}

As mentioned earlier, methods that take into account the data distribution can outperform \ac{MD} methods.
Following \cite{molchanov2019importance}, we define the \ac{DD} importance score by:

\begin{equation}
    \label{eqn:param_importance}
    \Importance{DD}(\AnyParameter{i}) =
    \Big ( \Likelihood(\Data, \AnyModel) - \Likelihood(\Data, \AnyModel | \AnyParameter{i}  = 0) \Big )^2 ,
\end{equation}
\noindent
where $\Likelihood(\cdot)$ is a training loss function, $\Data$ the data, $\AnyModel$ the model parameters and the condition $\AnyParameter{i}  = 0$ represents masking the parameter \AnyParameter{i}.

Equation~\ref{eqn:param_importance} is intractable as it requires a forward pass for each masked parameter.
Fortunately, we can use the first-order Taylor approximation, which correlates well with the true optimizer \cite{molchanov2019importance}:
\begin{equation}
    \label{eqn:param_importance_approx_taylor}
    \Importance{DD}(\AnyParameter{i}) \approx
    \Big ( \Gradient{i}\Likelihood(\Data, \AnyModel) \cdot \AnyParameter{i} \Big )^2 ,
\end{equation}
\noindent
where $\Gradient{i}\Likelihood(\cdot)$ is the local gradient of parameter $\AnyParameter{i}$.
Note that a single \editpostreview{forward}{forward-backward} pass is  \editpostreview{needed}{needed to compute importance scores} for all parameters.

\subsection{Pruning methods}
\label{sec:pruning_methods}

The \textit{pruning method} defines which components of the model should be removed at each pruning step, for example, individual weights, matrix rows or columns, or entire layers.

\subsubsection{Unstructured pruning}
\editpostreview{Unstructured pruning}{This method} introduces sparsity in the network by individually considering the model weights to remove.
Given a model $\AnyModel$, we can remove any connection to the weight \AnyParameter{i} without taking into account any parameter \AnyParameter{{j\ne i}}.

Pruning is implemented by applying the Hadamard product of the model parameters with a binary mask $M$, which has ones on the parameters to keep.
\editpostreview{}{We apply masking at training and inference to keep track of the masked weights.}

\subsubsection{Structured pruning}
\editpostreview{Structured pruning}{This method} considers groups of weights to remove altogether.
Given the model \AnyModel, each pruning step removes weights \AnyParameter{i \iff i \in G_k}, where
\editpostreview{$G_k \in \R^N$}{$G_k$} is the set of indices of group $k$ to remove.
Indices groups are disjoint, that is, a weight belongs to a single group.

We apply pruning by masking out rows or columns of layer projections.
Therefore, we use an extension of the unstructured pruning algorithm, with the additional structural constraint.
Again, pruning masks are used for training and inference.
Techniques such as matrix reordering can improve inference speed \cite{niu2022grim},
but this is not in the scope of this work.

The importance scores of rows or columns are computed by averaging the scores of their weights.
We also experimented using the minimum \editpostreview{function}{or the maximum}, and the average yielded better accuracy.

\subsubsection{Pruning factorized layers}
\label{par:factorized_layers}

Low-rank approximation \editpostreview{methods}{techniques} represent a matrix
as the product of smaller matrices, in such a way that the 
reconstruction loss is minimized.
Given a matrix $W \rightarrow \R^{a\times b}$, there exists a set of matrices
$U\rightarrow \R^{a\times r}$, $V\rightarrow \R^{r\times b}$ and a 
diagonal matrix $D\rightarrow \R^{r\times r}$ such as $W\approx UDV$.
In common \editpostreview{methods}{approaches}, like \editpostreview{}{the} ones based on \acl{SVD}, the number of parameters initially increases, given that $r = \min(a,b)$.
Then, pruning is applied based on the magnitude of the diagonal values of $D$, so that, usually, $r\ll \min(a,b)$.

\begin{figure}[t]
    \centering
    \includegraphics[width=7cm]{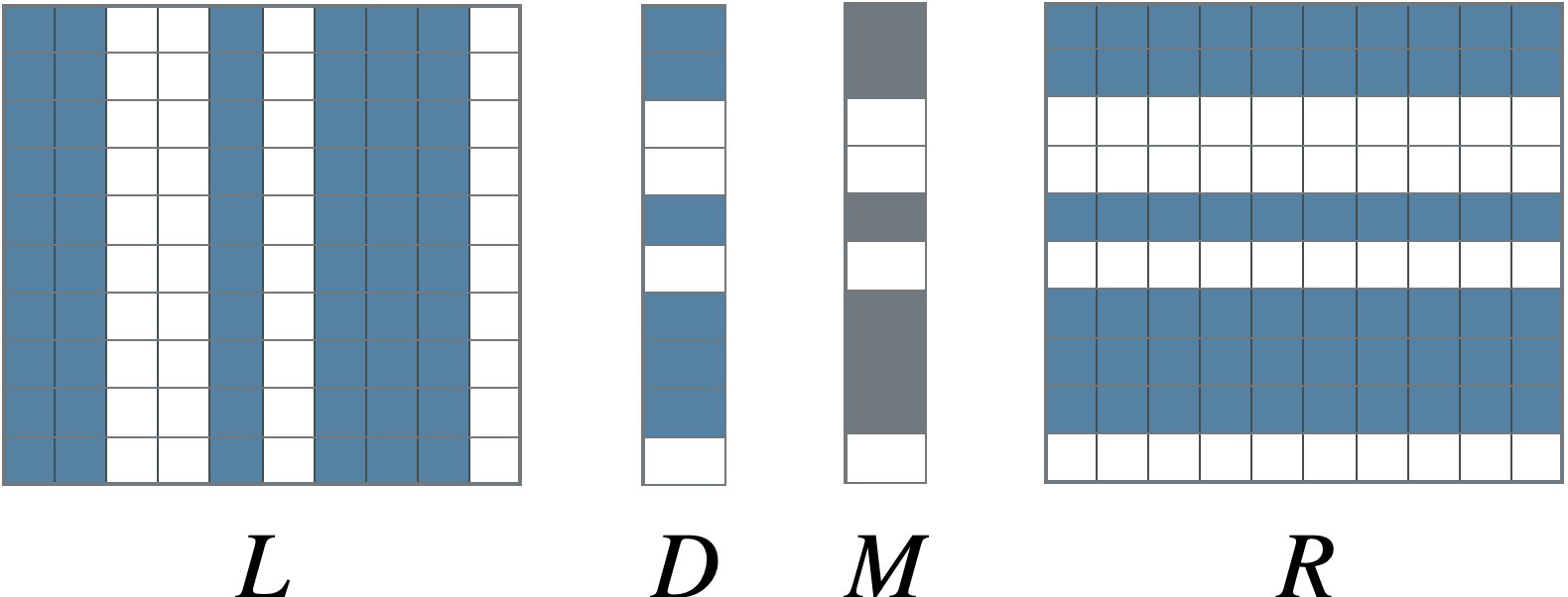}
    \caption{Representation of pruning factorized layers. Diagonal matrices are represented as diagonal vectors.}
    \label{fig:decomposition_algorithm}
\end{figure}

In this work, we build on this idea.
We first apply matrix factorization to the projection layers,
then, include a diagonal binary mask $M\rightarrow \R^{r\times r}$, such as the approximation becomes $W\approx UDMV$.
Pruning is applied via masking, using a similar algorithm as for unstructured pruning; see Figure~\ref{fig:decomposition_algorithm}.
This formulation allows a fine-grained control of pruning criteria, given that importance scores \editpostreview{}{can} depend on $U$ and $V$, in addition to $D$.
\editpostreview{
This can be especially useful for \acl{DD} pruning (Section~\ref{sec:pruning_scheduler}).
We apply $l_1$ regularization to encourage sparsity on the diagonal matrix, similar to Wang et al. (2020).}{}
\editpostreview{}{However, we obtained better results using diagonals only.}

\subsection{Pruning scheduler}
\label{sec:pruning_scheduler}

The \textit{scheduler} defines how many pruning steps to run and how many weights to remove in each step. In \textit{one-shot} pruning, the lowest-importance weights are all removed in a single step, with the number of weights defined by the target sparsity.

As shown in \cite{zhu2018toprune},
\textit{one-shot} does not allow the model to recover from pruning losses.
Following the authors, we formulate the \textit{incremental} polynomial
scheduler as follows.
Given the initial sparsity $s_i$, the final sparsity $s_f$, a pruning frequency $\Delta t$, and the number of pruning steps $n$, the target sparsity at time-step $t$ is denoted by:
\begin{equation}
\label{eqn:pruning_scheduler}
s_t = s_f + (s_i - s_f)  \left( 1 - \frac{t - t_0}{n \Delta t}  \right)^3
\end{equation}

The polynomial scheduler prunes more aggressively at the beginning of training and slowly converges to the final sparsity towards the end.
The intuition is that the initial network is over-parameterized and, as training converges, we  apply minor adjustments to let the model recover from pruning-induced losses.

The scheduler needs to be adapted for factorized layers to take into account that:
a) the initial decomposition increases the number of parameters;
and b) pruning a single value of $D$ entails the removal of an entire column and row from $U$ and $V$.
Hence, we propose an updated target sparsity at step $t$.
Given a matrix $W \rightarrow \R ^{a \times b}$, we compute:

\begin{equation}
\label{eqn:pruning_scheduler_updated}
\hat{s}_t = 1 - a * \frac{1-s_t}{a+b} .
\end{equation}

\section{Experiments}

\subsection{Experimental setup}
\label{ssec:exp_setup}

\subsubsection{Language models}
The \acp{NNLM} used in this work are based on Transformers \cite{vaswani2017attention}.
The input layer encodes 6300 subword units,
obtained with the \emph{sentencepiece} algorithm \cite{kudo2018sentencepiece},
into 512-dimensional embedding vectors.
Relative positional encoding is added \cite{shaw-etal-2018-self-short} to the input, followed by
six stacked encoder blocks similar to \cite{zhang2020transformer}.
Our architecture uses layer normalization, followed by a causal multi-head attention block with eight 128-dimensional heads and forward residual connections. The final block is a linear projection with softmax activation.

\editpostreview{
The baseline \emph{standard} and \emph{small} models have about 20.6M and 10.2M parameters respectively.
In order to achieve the desired size of \emph{small} and \emph{pruned} models, the dimensions of layer projections is reduced.   
Model pruning starts from a checkpoint taken after 50 training epochs; each epoch has 8k steps.
}
{
The baseline model has about 20M parameters,
and model pruning starts from a checkpoint taken after 50 epochs; each epoch has 8k training steps.
Additional smaller models were trained from scratch for comparison, with the target sizes
achieved by reducing the dimensions of layer projections and keeping the \editpostreview{}{same} layer
arrangement\editpostreview{of the baseline model}{}.
}

To avoid removal of entire subwords due to the sparsity of rows in the gradients of the embedding matrix~\cite{anonymous2023decepticons}, we \editpostreview{aggregate}{group} importance scores across the vocabulary dimension.

\subsubsection{Speech recognition system}
The speech recognition experiments use a \ac{TT} model \cite{zhang2020transformer} with 102M parameters in total.
The architecture is similar to \cite{swietojanski2022variable},
with a 12-layer Conformer-based acoustic encoder \cite{gulati2020conformer} and a six-layer Transformer-based label encoder.
The acoustic and label encodings are summed and fed into the joint network, which has a single projection to the 6300 subword units, and softmax activation.
In this work, the \ac{TT} model is fixed. Note it could be possible to prune the \ac{TT} layers as well, given it has similar components as the \acp{NNLM}.

\editpostsubmission{}{The \ac{TT} model was trained on the transducer loss~\cite{graves2012transducer} for 90 million steps. Distributed synchronized gradient-descent with exponential moving averaging was used~\cite{keskar2017largebatch}. The input  are 80-dimensional filter-bank acoustic vectors with SpecAugment~\cite{park2019specaugment}.}

Decoding is carried out using beam search, with an external \ac{LM} shallow fusion \cite{zhao2019shallow-short} and internal \ac{LM} subtraction \cite{mcdermott2019,meng2021}.
\editpostsubmission{}{The internal and external \ac{LM} weights were optimized for each decoding experiment.}

\subsubsection{Data}
We conduct experiments using a subset of English text data containing about 1.7B utterances
from dictation and \ac{VA} tasks.
The vast majority comprises automatic, weakly supervised and manual transcriptions
of anonymized requests randomly sampled from opt-in users.
The automatic transcriptions are generated by a distinct \ac{ASR} recognizer.
A small part of the data, about 6\%, consists of utterances synthetically generated from
domain-specific templates, similar to \cite{vangysel2022}, where the templates are also
derived from anonymized data randomly sampled from opt-in users.

\subsubsection{Evaluation metrics}
We use accuracy and performance metrics to assess the impact of pruning.
Data \ac{PPL} is computed for each \ac{NNLM} on a 50k-utterance development set covering dictation and \ac{VA} tasks.
\Ac{NNLM} \editpostsubmission{performance}{complexity} is measured using the estimated number of \acp{FLOP} required for inference.
We report speed-ups as the \ac{FLOP} ratio between the baseline and \editpostreview{}{the} pruned models.

\Ac{ASR} decoding accuracy is reported in terms of \ac{WER}\editpostsubmission{, 
and \ac{SNEER}, which measures whether a named-entity has been fully recognized or not.}{.}
\ac{ASR} evaluation sets cover the dictation and \ac{VA} tasks, with about 23k and 49k requests respectively.

\subsection{Language modeling results}
\label{ssec:lm_results}

We first analyze the contribution of the three aspects of pruning in terms of data \acl{PPL} and inference performance.
Unless stated otherwise, the default pruning scheme uses the \emph{magnitude} criterion, the \emph{unstructured}
method\editpostreview{and}{,} the \emph{one-shot} scheduler\editpostreview{}{, and equally applies to all the layers in the network.}
\editpostsubmission{A}{We analyze individual layer contributions to model pruning and report the results in  \editpostsubmission{Appendix}{Section}~\ref{section:layer_analysis}.}

\subsubsection{Pruning schedulers}

A comparison of the pruning schedulers (Section~\ref{sec:pruning_scheduler}) is shown in Table~\ref{tab:ppl_incremental_vs_oneshot}.
Incremental pruning achieves better results for all target sizes, what
aligns with observations from \cite{zhu2018toprune}.
In addition, we observe that incremental pruning is crucial for high compression rates.
Notably, with 5\% of the model size (1M), incremental pruning outperforms one-shot by 64.8\% relative. 
We note, however, that one-shot pruning is competitive for moderate compression rates.
With 50\% of the model size (10M), the \ac{PPL} difference is about 1.8\% relative.
Similar to \cite{xu2022importance}, we attribute the small difference to the fact
the initial model is over-parametrized.

\begin{table}[t]
    \centering
    \begin{tabular}{cccr}
    \toprule
    \multirow{2}{*}{\textbf{Model}} & \multicolumn{3}{c}{\textbf{\Acl{PPL}}} \\
    \cmidrule(l){2-4}
    \textbf{size} & \textbf{One-shot} & \textbf{Incremental} & $\Delta \%$\\
    \midrule
    10M & 16.5 & \textbf{16.2} & -1.8  \\
     5M & 19.8 & \textbf{17.2} & -12.2  \\
     2M & 33.1 & \textbf{19.6} & -40.8 \\
     1M & 64.0 & \textbf{22.5} & -64.8  \\
    \bottomrule
    \end{tabular}
    \caption{
    \editpostsubmission{\Acl{PPL} of the development data, comparing pruning \emph{schedulers}.}{Comparison of pruning \emph{schedulers}.}
    The \ac{PPL} with the baseline 20M \ac{NNLM} is 16.0.
    $\Delta\%$ is the relative change in \ac{PPL}.
    \textbf{Bold} scores indicate the best results.
    }
    \label{tab:ppl_incremental_vs_oneshot}
\end{table}

\begin{figure}[t]
    \centering
    \includegraphics[width=8cm]{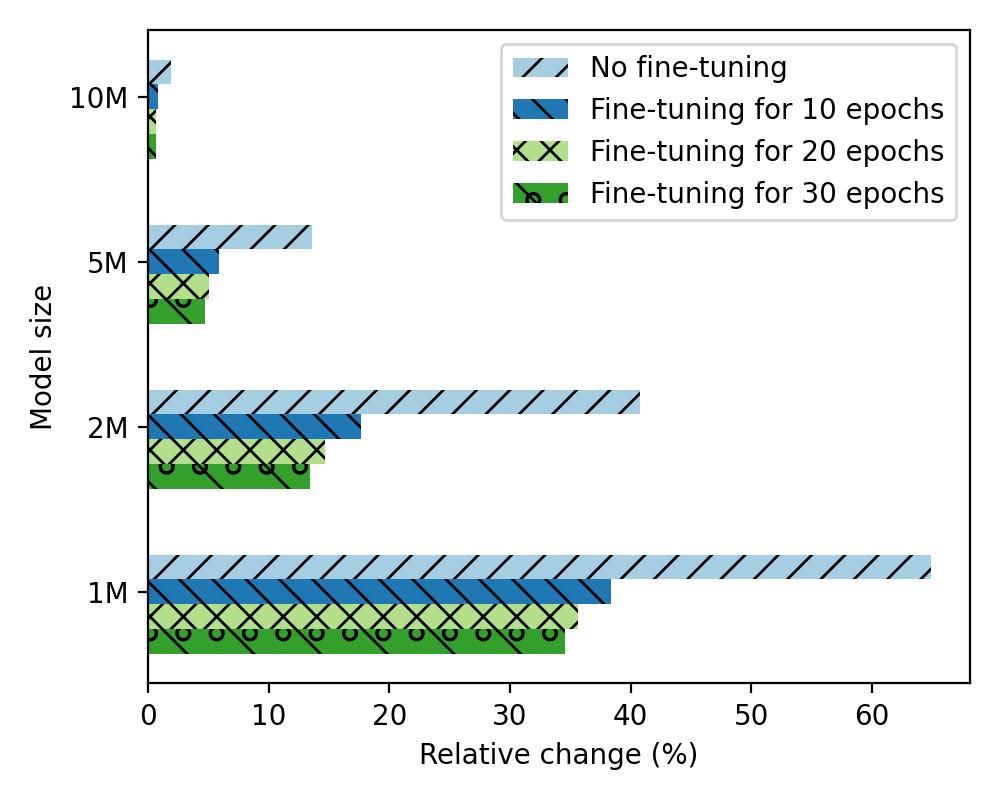}
    \caption{
    \editpostsubmission{
    Relative \acl{PPL} change between incremental and one-shot pruning for different target model sizes.
    }{
    Relative \acl{PPL} change of incremental pruning compared to one-shot pruning for various target model sizes.
    }
    }
    \label{fig:model_finetuning}
\end{figure}

Next, we conducted a set of experiments to isolate the effect of fine-tuning.
We first apply pruning up to the target size using either pruning schedulers, then continue with regular model training.
The results are shown in Figure~\ref{fig:model_finetuning}.
As expected, the gap between one-shot and incremental pruning reduces
as the fine-tuning process continues.
Nevertheless, the incrementally-pruned model still obtains better results even after 30 epochs.
With a 1M-parameter target size, incremental pruning outperforms one-shot pruned fine-tuned models by 35\% relative.
This highlights that wrong pruning decisions cannot be recovered with post tuning.

\subsubsection{Pruning criteria}
\label{ssec:exp_pruning_criteria}

A comparison of pruning criteria (Section~\ref{sec:pruning_criteria}) is shown in Table~\ref{tab:ppl_mb_vs_ib}.
We report results with unstructured incremental pruning, and observed similar trends with other setups.

Generally, \acl{DD} outperforms \acl{MD} pruning, as noted by \cite{molchanov2019importance}.
In addition to prior work, we observe that the gap between \ac{DD} and \ac{MD} becomes larger as the target size reduces.
With 5\% of the model size (1M), \ac{DD} pruning outperforms \ac{MD} pruning by 1.8\% relative.

\begin{table}[t]
    \centering
    \begin{tabular}{cccr}
    \toprule
    \multirow{2}{*}{\textbf{Model}} & \multicolumn{3}{c}{\textbf{\Acl{PPL}}} \\
    \cmidrule(l){2-4}
    \textbf{size} & \textbf{Magnitude} & \textbf{Data} & $\Delta \%$\\
    \midrule
    10M & 16.2 & \textbf{16.2} & 0.0 \\
     5M & 17.2 & \textbf{17.0} & -1.2 \\
     2M & 19.6 & \textbf{19.3} & -1.5 \\
     1M & 22.5 & \textbf{22.1} & -1.8 \\
    \bottomrule
    \end{tabular}
    \caption{
    \editpostsubmission{\Acl{PPL} of the development data, comparing pruning \emph{criteria}.}{Comparison of pruning \emph{criteria}.}
    The \ac{PPL} with the baseline 20M \ac{NNLM} is 16.0.
    $\Delta\%$ is the relative change in \ac{PPL}.
    \textbf{Bold} scores indicate the best results.
    }
    \label{tab:ppl_mb_vs_ib}
\end{table}

\subsubsection{Pruning methods}
A comparison of pruning methods (Section~\ref{sec:pruning_methods}) is shown in
Table~\ref{tab:ppl_methods_comparison}.
We report results with \acl{DD} incremental pruning, and observed similar outcome
with \acl{MD} pruning.

\emph{Unstructured} pruning yields the best \acl{PPL},
outperforming the other pruning methods, as well as
models trained \emph{from scratch} with the same size.
In addition, we note the unstructured pruned 10M-parameter model \editpostsubmission{is only 1.2\% relative behind the 20M-parameter baseline.}{is comparable to the 20M-parameter baseline, lagging only 1.2\% relative behind.}

\emph{Structured} pruning falls largely behind.
We speculate the degradation is due to the inherent binding between connecting layers:
removing a column of a layer entails the removal of a row of following neighbor layers,
including  residual ones.
We assessed pruning rows or columns of hidden projections, and obtained similar results.

The \emph{factorized} method does not suffer from this issue.
For a moderate compression rate (10M), the proposed factorized method obtains a \acl{PPL} only 2.7\% behind the model trained from scratch.
However, the gap is higher (13.6\% relative) for smaller sizes (1M).
We conjecture this is due to drastic reductions of the inner dimensionality of factorized
layers.
For example, pruning a $512$-dimensional square matrix to 5\% of its size
means factorizing it into a $512\times 12$ projection followed by a \editpostsubmission{$512\times 12$}{$12\times 512$} one.

With our implementation, only the \emph{factorized} method yields inference \editpostreview{speed}{speed-ups}.
With this setup, the 10M model is $1.8\times$ faster than the baseline 20M model, and $1.1\times$ slower than the 10M model trained from scratch.
Considering the accuracy and speed results observed, the factorized method may suit well pipelines that aim at generating models to fit into multiple resources constraints,
given that incrementally pruning a model is generally faster than training multiple models from scratch.
However, for strong pruning rates, training models from scratch may\editpostsubmission{}{still} be a better option.

\begin{table}[t]
    \centering
    \begin{tabular}{rrrrr}
    \toprule
    \textbf{NNLM} & \textbf{10M} & \textbf{5M} & \textbf{2M} & \textbf{1M} \\
    \midrule
    from scratch       & 17.1 & 18.5 & 21.3 & 24.6 \\
    \midrule
    unstructured    & \textbf{16.2} & \textbf{17.0} & \textbf{19.3} & \textbf{22.1} \\
    \hspace{4em}$\mathit\Delta\%$ &
        \reldiff{-5.3} & \reldiff{-7.8} & \reldiff{-9.5} & \reldiff{-10.4} \\
    \midrule    
    structured      & 19.8 & 26.3 & 41.9 & 60.6 \\
    \hspace{4em}$\mathit\Delta\%$ &
        \reldiff{15.7} & \reldiff{42.4} & \reldiff{96.4} & \reldiff{146.2} \\
    \midrule
    factorized  & 17.6 & 19.2 & 22.9 & 28.0 \\
    \hspace{4em}$\mathit\Delta\%$ &
        \reldiff{2.7} & \reldiff{4.2} & \reldiff{7.4} & \reldiff{13.6} \\
    \bottomrule
    \end{tabular}
    \caption{
    \editpostsubmission{\Acl{PPL} of the development data, comparing pruning \emph{methods}.}{Comparison of pruning \emph{methods}.}
    The \ac{PPL} with the baseline 20M \ac{NNLM} is 16.0.
    $\Delta\%$ is the relative change in \ac{PPL} compared to models
    trained \emph{from scratch} with the same size (in the first row).
    \textbf{Bold} scores indicate the best results.
    }
    \label{tab:ppl_methods_comparison}
\end{table}

\subsubsection{Layer-by-layer analysis}\label{section:layer_analysis}
\begin{figure}[t]
    \centering
    \includegraphics[width=8cm]{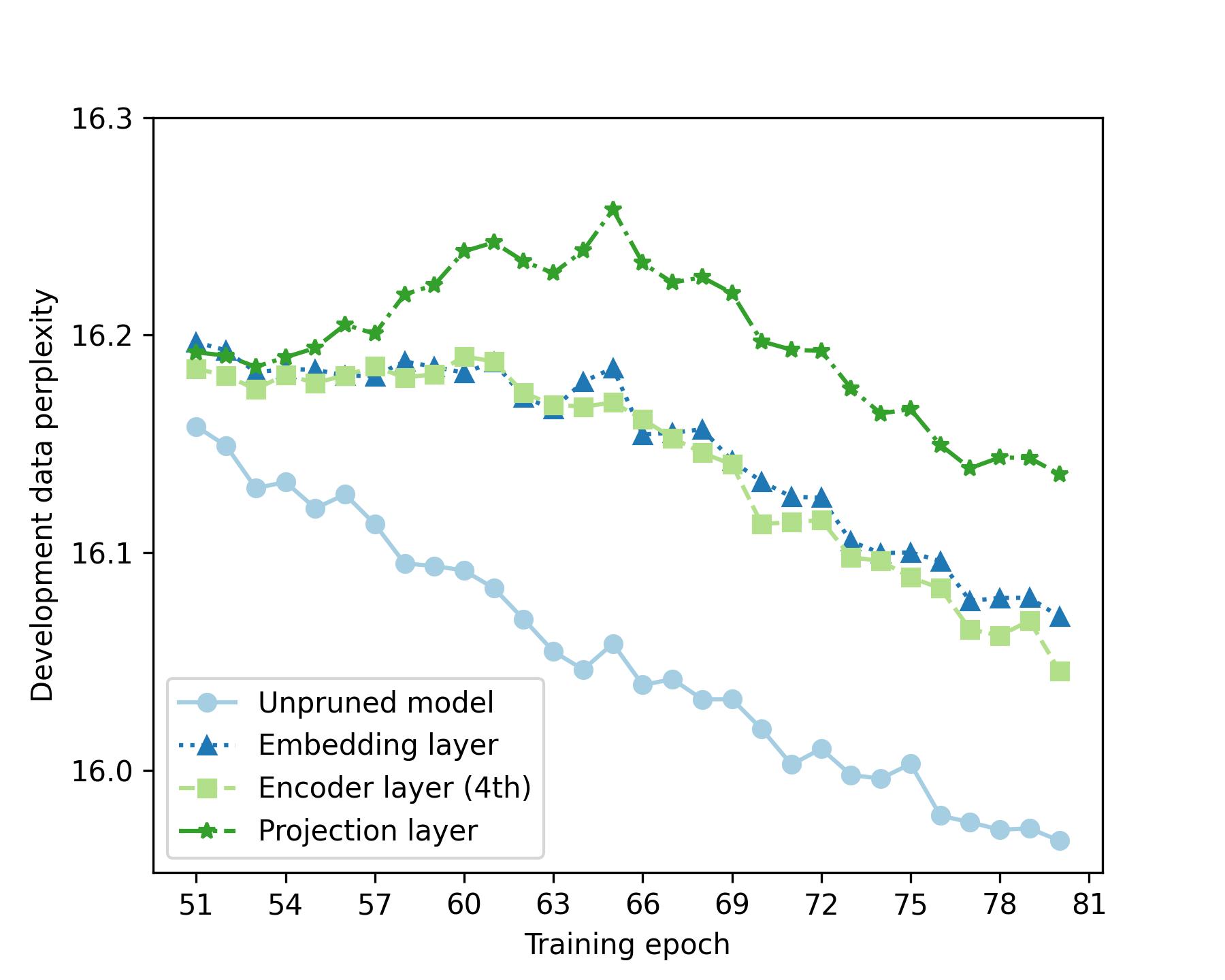}
    \caption{
    \editpostreview{}{Comparison of pruning individual layers of the 20M \ac{NNLM}.
    All the pruned models have 18.4M or 19M parameters.
    \editpostreview{\Acf{PPL} of the combined dictation and \acl{VA} development data.}{}
    \editpostsubmission{}{\ac{PPL} increases until the network becomes robust to pruning and reaches a peak around the 65th training epoch (15th pruning epoch).}}
    }
    \label{fig:comparison_75_fine_grained}
\end{figure}

To understand whether certain layers are more affected than others, we prune individual layers to 75\% sparsity with unstructured \ac{MD} pruning and \editpostreview{compare}{report} the results \editpostreview{}{in Figure~\ref{fig:comparison_75_fine_grained}}.
We report results for an intermediate encoder layer to avoid visual clutter; the metrics are close for the other encoder layers.

Pruning the output projection layer results in the highest \ac{PPL} degradation,
while pruning the encoder or the embedding layer are equivalent in terms of \ac{PPL}.
\editpostsubmission{}{We attribute it to the fact top layers are task-specific and as such more sensible to sudden changes in the norm of the weights, in turn making them brittle to weights pruning.}

\editpostsubmission{Furthermore}{Moreover}, \ac{PPL} degradation is limited to 2\% when pruning the attention heads.
\editpostsubmission{}{It is a remarkable finding to further compress the model size without compromising accuracy as they amount to the 55\% of all the encoder parameters.}

\subsection{\Acl{ASR} results}
\label{ssec:asr_results}

We conducted additional experiments to evaluate how the pruning aspects contribute to the accuracy
of \acl{ASR} where the \ac{NNLM} is used in shallow fusion beam decoding.
Generally, the trends are similar to those discussed in Section~\ref{ssec:lm_results}.
We did not observe significant \ac{WER} differences while comparing the pruning \emph{criteria}.
In addition, \emph{incremental} pruning consistently outperforms \emph{one-shot},\editpostsubmission{especially
for smaller target sizes}{closely following the trends reported for perplexity but with smaller gaps given the acoustic \ac{TT} model is fixed,}\editpostsubmission{. Such results are reported in Appendix B}{as shown in Table~\ref{tab:wer_incremental_vs_oneshot_app}}.
\editpostsubmission{}{
The majority of the results is statistically significant according to the so-called \ac{MAPSSWE} test~\cite{gillick1989stat} with $p<0.005$.}

\begin{table}[t]
    \centering
    \begin{tabular}{cccr}
    \toprule
    \multirow{2}{*}{\textbf{Model}} & \multicolumn{3}{c}{\textbf{\Acl{WER}}} \\
    \cmidrule(l){2-4}
    \textbf{size} & \textbf{One-shot} & \textbf{Incremental} & $\Delta \%$\\
    \midrule
    10M & 3.49 & \textbf{3.48} & -0.3$^*$\\
    5M  & 3.57  & \textbf{3.51} & -1.7$\phantom{*}$ \\
    2M  & 3.70  & \textbf{3.62} & -2.2$\phantom{*}$ \\
    1M  & 3.78  & \textbf{3.69} & -2.4$\phantom{*}$ \\
    \midrule
    \midrule
    10M & 4.22  & \textbf{4.21} & -0.2$^*$ \\
    5M  & 4.31  & \textbf{4.28} & -0.7$^*$ \\
    2M  & 4.44  & \textbf{4.35} & -2.0$\phantom{*}$ \\
    1M  & 4.57  & \textbf{4.43} & -3.1$\phantom{*}$ \\
    \bottomrule
    \end{tabular}
    \caption{
    Comparison of pruning \emph{schedulers} on dictation (upper) and \acf{VA} (lower) evaluation data.
    The baseline \ac{WER} for the 20M \ac{NNLM} is 3.47\% on dictation and 4.24\% on \ac{VA}.
    $\Delta\%$ is the relative change in \ac{WER}.
    \textbf{Bold} scores indicate the best results.
    \editpostsubmission{}{
    The star ($^*$) denotes changes that are \emph{not} statistically significant ($p>0.005$).}
    }
    \label{tab:wer_incremental_vs_oneshot_app}
\end{table}

\begin{table}[t]
    \centering
    \begin{tabular}{lcc}
    \toprule
    \multirow{2}{*}{\textbf{NNLM}} & \multicolumn{2}{c}{\textbf{\Acl{WER}}} \\
    \cmidrule(l){2-3}
    & \textbf{Dictation} & \textbf{\Acl{VA}} \\
    \midrule
    20M-scratch & 3.47\phantom{*} & 4.24$\phantom{*}$ \\
    10M-scratch & 3.53\phantom{*} & 4.27$\phantom{*}$ \\
    \midrule
    unstructured & \textbf{3.48}\phantom{*} & \textbf{4.21}$\phantom{*}$ \\
    structured   & 4.21\phantom{*} & 5.26$\phantom{*}$ \\
    factorized   & 3.54$^*$ & 4.29$^*$ \\   
    \midrule
    \editpostreview{none}{no \ac{NNLM}}         & 3.77\phantom{*} & 4.47$\phantom{*}$ \\  
    \bottomrule
    \end{tabular}
    \caption{
    Comparison of pruning \emph{methods} with \acp{NNLM} trained from scratch (upper), pruned to 10M parameters (middle), and without \editpostsubmission{}{shallow fusion} (lower).
    \textbf{Bold} scores indicate the best results.
    \editpostsubmission{}{
    The star ($^*$) denotes results that are \emph{not} statistically significant ($p>0.005$) compared to \emph{10M-scratch}.}
    }
    \label{tab:wer_methods_comparison}
\end{table}

Table~\ref{tab:wer_methods_comparison} shows a comparison of pruning \emph{methods} using
incremental \ac{DD} pruning, \editpostsubmission{}{as the combination yielded the best results}.
\emph{Unstructured} pruning obtains better \ac{ASR} accuracy compared to other
pruning methods.
The \ac{WER} differences are smaller than the \ac{PPL} ones, which
is expected, given that the 102M-parameter \acl{TT} is fixed.
We leave the investigation of \ac{TT} pruning for future work.
The last row shows the results without external \acp{NNLM}.

\subsubsection{Data distribution}
We further analyzed the \ac{ASR} results
\editpostsubmission{as a function of}{based}
on the data distribution.
Table~\ref{tab:rare_entities_analysis_mb} reports the relative \ac{WER}
\editpostsubmission{difference}{change}
between a pruned \ac{NNLM} and one trained from scratch, both with 10M parameters.
\editpostsubmission{results reported use}{Reported results use}
unstructured incremental \ac{DD} pruning,
\editpostsubmission{and we observed similar trends with other setups.}{with similar trends observed for other setups,}
\editpostsubmission{The percentiles are obtained as a function of the word frequencies in the requests}{ and the percentiles represent word frequencies in the requests.}

The results suggest that pruned models are able to keep their accuracy on less frequent data (75\% and 100\% percentiles).
It is particularly interesting considering that \acp{NNLM} play an important role in the recognition of 
less frequent data in all-neural \ac{ASR} systems as the one used in this work.

\begin{table}[t]
    \centering
    \begin{tabular}{lcc}
    \toprule
    \multirow{2}{*}{\textbf{Percentile}} & \multicolumn{2}{c}{\textbf{\Acl{WER} relative}} \\
    \cmidrule(l){2-3}
    & \textbf{Dictation} & \textbf{\Acl{VA}} \\
    \midrule
    25\%  & -1.3 &  \phantom{-}0.0 \\
    50\%  & -0.7 &  \phantom{-}1.3 \\
    75\%  & \textbf{-2.6} & \textbf{-2.0} \\
    100\% & -1.2 & -1.9 \\
    \bottomrule
    \end{tabular}
    \caption{
    Impact of pruning on data percentiles.
    Each percentile has 25\% of words, from the most (25\%) to the least (100\%)
    frequent utterances.
    \editpostreview{We report the relative \ac{WER} and \ac{SNEER} changes between
    the \acl{DD} unstructured incremental pruned and trained from scratch 10M models.}{}
    \textbf{Bold} indicates the best results.
    }
    \label{tab:rare_entities_analysis_mb}
\end{table}

\section{Conclusions}
In this paper, we devise a comprehensive benchmark on model pruning and analyze the
\editpostsubmission{key}{primary}
issues
\editpostsubmission{of}{involved in}
pruning Transformer-based \acp{NNLM} for \ac{ASR}.

Incremental pruning enables high compression rates with limited accuracy degradation, and outperforms one-shot pruning up to 65\% relative \ac{PPL} on the smallest target size
assessed.
We showed that additional fine-tuning is not enough to close the gap to the incremental scheduler.
On the other hand, one-shot pruning can be an efficient training-free alternative for moderate compression rates, with negligible accuracy degradation.
We found that \acl{DD} pruning consistently outperforms \acl{MD} pruning up to 2\% relative \ac{PPL}.

Unstructured-pruned models outperform the same-size models trained from scratch on
various target sizes.
In particular, the 10M pruned \ac{NNLM} yields 1\% \ac{WER} relative improvement
on dictation and \ac{VA} data over the 10M \ac{NNLM} trained from scratch.
Interestingly, most of such gains were observed on less frequent data.

Pruning factorized layers achieves comparable \ac{WER} to the model trained from scratch,
and a reasonable speed-up compared to the baseline.
Therefore, it could serve as a setup to generate multiple small models incrementally
for moderate compression rates.
For extremely high compression rates, training from scratch may be a suitable
alternative to trade off accuracy and performance.

\section{ACKNOWLEDGMENTS}
\label{sec:ack}
We would like to thank Arturo Argueta, Amr Mousa, Lyan Verwimp, Mirko Hanemann, Caglar Tirkaz and Manos Tsagkias for their inputs and useful discussion.

\end{document}